\pdfoutput=1

\documentclass[11pt]{article}

\usepackage{acl}

\usepackage{times}
\usepackage{latexsym}

\usepackage[T1]{fontenc}

\usepackage[utf8]{inputenc}

\usepackage{microtype}

\usepackage{algorithm}
\usepackage{algpseudocode}
\usepackage{amsmath}
\usepackage{amssymb}
\usepackage{amsmath}
\usepackage{arydshln}
\usepackage{ascii}
\usepackage{bm}
\usepackage{booktabs}
\usepackage{colortbl}
\usepackage{color}
\usepackage{CJK}
\usepackage{dsfont}
\usepackage{enumitem}
\usepackage{forest}
\usepackage{graphics}
\usepackage{graphicx}
\usepackage{latexsym}
\usepackage{multirow}
\usepackage{mwe}
\usepackage{makecell}
\usepackage{microtype}
\usepackage{subfig}
\usepackage{times}
\usepackage{tikz}
\usepackage{tabularx}
\usepackage{url}
\usepackage{wrapfig}

\definecolor{sred}{RGB}{196, 38, 11}
\definecolor{sblue}{RGB}{41, 52, 190}
\definecolor{sgreen}{RGB}{18, 141, 21}

\title{Integrating Vectorized Lexical Constraints
\\ for Neural Machine Translation}

\author{
    Shuo Wang$^1$ \quad Zhixing Tan$^1$ \quad Yang Liu$^{1,2,3,4,5}$\thanks{\ \ Correspondence to: Yang Liu.} \\
    $^1$Dept. of Comp. Sci. \& Tech., Institute for AI, Tsinghua University, Beijing, China \\
    Beijing National Research Center for Information Science and Technology \\
    $^2$Institute for AI Industry Research, Tsinghua University, Beijing, China \\
    $^3$International Innovation Center of Tsinghua University, Shanghai, China \\
    $^4$Quan Cheng Laboratory \ $^5$Institute for Guo Qiang, Tsinghua University, Beijing, China\\
    \href{mailto:wangshuo.thu@gmail.com}{\asciifamily{wangshuo.thu@gmail.com}} \quad \href{mailto:zxtan@tsinghua.edu.cn}{\asciifamily{\{zxtan, liuyang2011\}@tsinghua.edu.cn}}\\
}

\begin{document}
\maketitle
\begin{abstract}

Lexically constrained neural machine translation (NMT), which controls the generation of NMT models with pre-specified constraints, is important in many practical scenarios. Due to the representation gap between discrete constraints and continuous vectors in NMT models, most existing works choose to construct synthetic data or modify the decoding algorithm to impose lexical constraints, treating the NMT model as a black box. In this work, we propose to open this black box by directly integrating the constraints into NMT models. Specifically, we vectorize source and target constraints into continuous keys and values, which can be utilized by the attention modules of NMT models. The proposed integration method is based on the assumption that the correspondence between keys and values in attention modules is naturally suitable for modeling constraint pairs. Experimental results show that our method consistently outperforms several representative baselines on four language pairs, demonstrating the superiority of integrating vectorized lexical constraints.~\footnote{For the source code, please refer to \url{https://github.com/shuo-git/VecConstNMT}.}

\end{abstract}

\section{Introduction}

Controlling the lexical choice of the translation is important in a wide range of settings, such as interactive machine translation~\cite{Koehn:2009:Lexical}, entity translation~\cite{Li:2018:Lexical}, and translation in safety-critical domains~\cite{Wang:2020:InfECE}. However, different from the case of statistical machine translation~\cite{Koehn:2007:MosesOS}, it is non-trivial to directly integrate discrete lexical constraints into neural machine translation (NMT) models~\cite{Bahdanau:2015:RNNSearch,Vaswani:2017:Transformer}, whose hidden states are all continuous vectors that are difficult for humans to understand.

In accordance with this problem, one branch of studies directs its attention to designing advanced decoding algorithms~\cite{Hokamp:2017:Lexical,Hasler:2018:Lexical,Post:2018:Lexical} to impose hard constraints and leave NMT models unchanged. For instance, \citet{Hu:2019:Lexical} propose a {\em vectorized dynamic beam allocation} (VDBA) algorithm, which devotes part of the beam to candidates that have met some constraints. Although this kind of method can guarantee the presence of target constraints in the output, they are found to potentially result in poor translation quality~\cite{Chen:2021:Lexical,Zhang:2021:Lexical}, such as repeated translation or source phrase omission.

Another branch of works proposes to learn constraint-aware NMT models through data augmentation. They construct synthetic data by replacing source constraints with their target-language correspondents~\cite{Song:2019:Lexical} or appending target constraints right after the corresponding source phrases~\cite{Dinu:2019:Lexical}. During inference, the input sentence is edited in advance and then provided to the NMT model. The major drawback of data augmentation based methods is that they may suffer from a low success rate of generating target constraints in some cases, indicating that only adjusting the training data is sub-optimal for lexical constrained translation~\cite{Chen:2021:Lexical}.

\begin{figure}[t]
    \centering
    \includegraphics[width=0.45\textwidth]{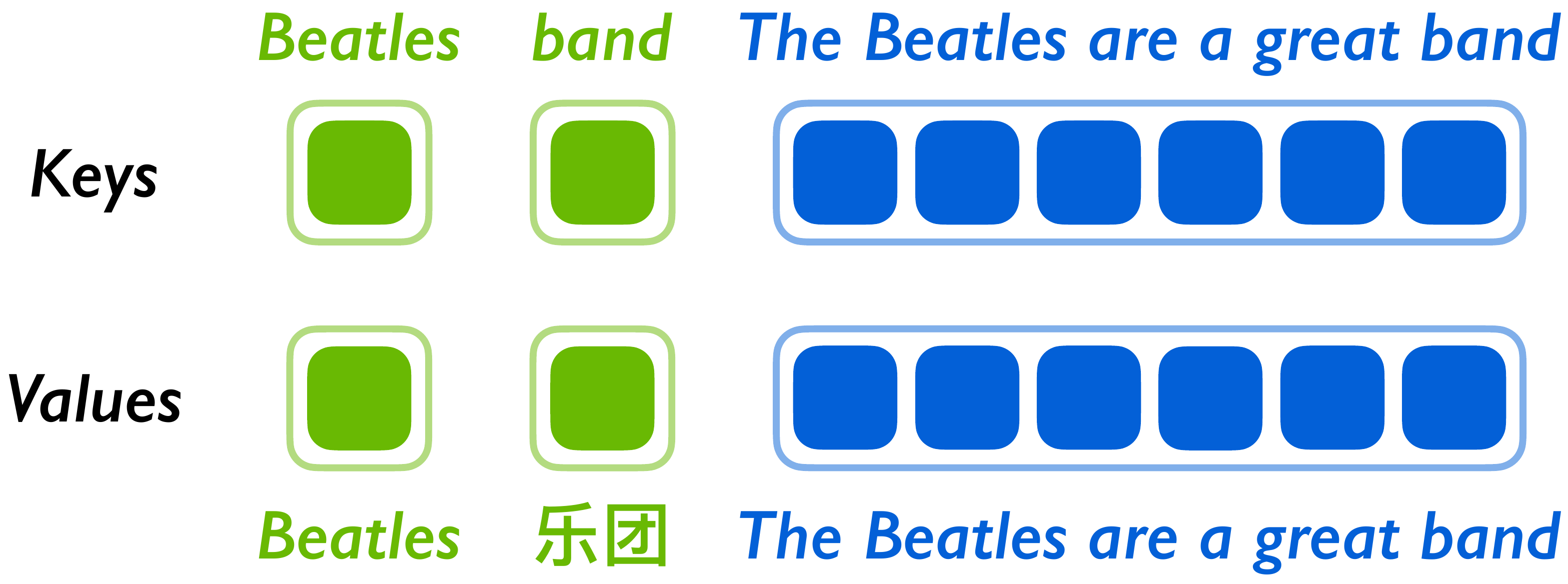}
    \caption{
        An example of the integration of vectorized lexical constraints into attention proposed in this work. We omit queries for simplicity. Blue and green squares denote the continuous representation of the source sentence and the constraints, respectively. The provided constraints are "Beatles$\rightarrow$Beatles" and "band$\rightarrow$\begin{CJK}{UTF8}{gbsn}乐团\end{CJK}".
    }
\label{fig:prompt-example}
\end{figure}

To make NMT models better learn from and cope with lexical constraints, we propose to leverage attention modules~\cite{Bahdanau:2015:RNNSearch} to explicitly integrate vectorized lexical constraints. As illustrated in Figure~\ref{fig:prompt-example}, we use vectorized source constraints as additional keys and vectorized target constraints as additional values.
Intuitively, the additional keys are used to estimate the relevance between the current query and the source phrases while the additional values are used to integrate the information of the target phrases.
In this way, each revised attention is aware of the guidance to translate {\em which source phrase} into {\em what target phrase}.

Experiments show that our method can significantly improve the ability of NMT models to translate with constraints, indicating that the correspondence between attention keys and values is suitable for modeling constraint pairs.
Inspired by recent progress in controlled text generation~\cite{Dathathri:2020:Plug,Pascual:2021:PlugAndPlay}, we also introduce a plug-in to the output layer that can further improve the success rate of generating constrained tokens. We conduct experiments on four language pairs and find that our model can consistently outperform several representative baselines.

\section{Neural Machine Translation}
\label{sec:nmt}
\paragraph{Training} The goal of machine translation is to translate a source-language sentence $\mathbf{x}=x_1 \dots x_{|\mathbf{x}|}$ into a target-language sentence $\mathbf{y}=y_1 \dots y_{|\mathbf{y}|}$. We use $P(\mathbf{y}|\mathbf{x}; \bm{\theta})$ to denote an NMT model~\cite{Vaswani:2017:Transformer} parameterized by $\bm{\theta}$. Modern NMT models are usually trained by maximum likelihood estimation~\cite{Bahdanau:2015:RNNSearch,Vaswani:2017:Transformer}, where the log-likelihood is defined as
\begin{equation}
    \log P(\mathbf{y}|\mathbf{x}; \bm{\theta}) = \sum_{t=1}^{|\mathbf{y}|} \log P(y_t|\mathbf{y}_{<t}, \mathbf{x}; \bm{\theta}),
\end{equation}
in which $\mathbf{y}_{<t}$ is a partial translation.

\paragraph{Inference} The inference of NMT models can be divided into two sub-processes:
\begin{itemize}
    \item {\em probability estimation}: the model estimates the token-level probability distribution for each partial hypothesis within the beam;
    \item {\em candidate selection}: the decoding algorithm selects some candidates based on the probability estimated by the NMT model.
\end{itemize}
These two sub-processes are performed alternatively until reaching the maximum length or generating the end-of-sentence token.

\begin{figure*}[t]
    \centering
    \subfloat[Overview]{\includegraphics[width=0.31\textwidth]{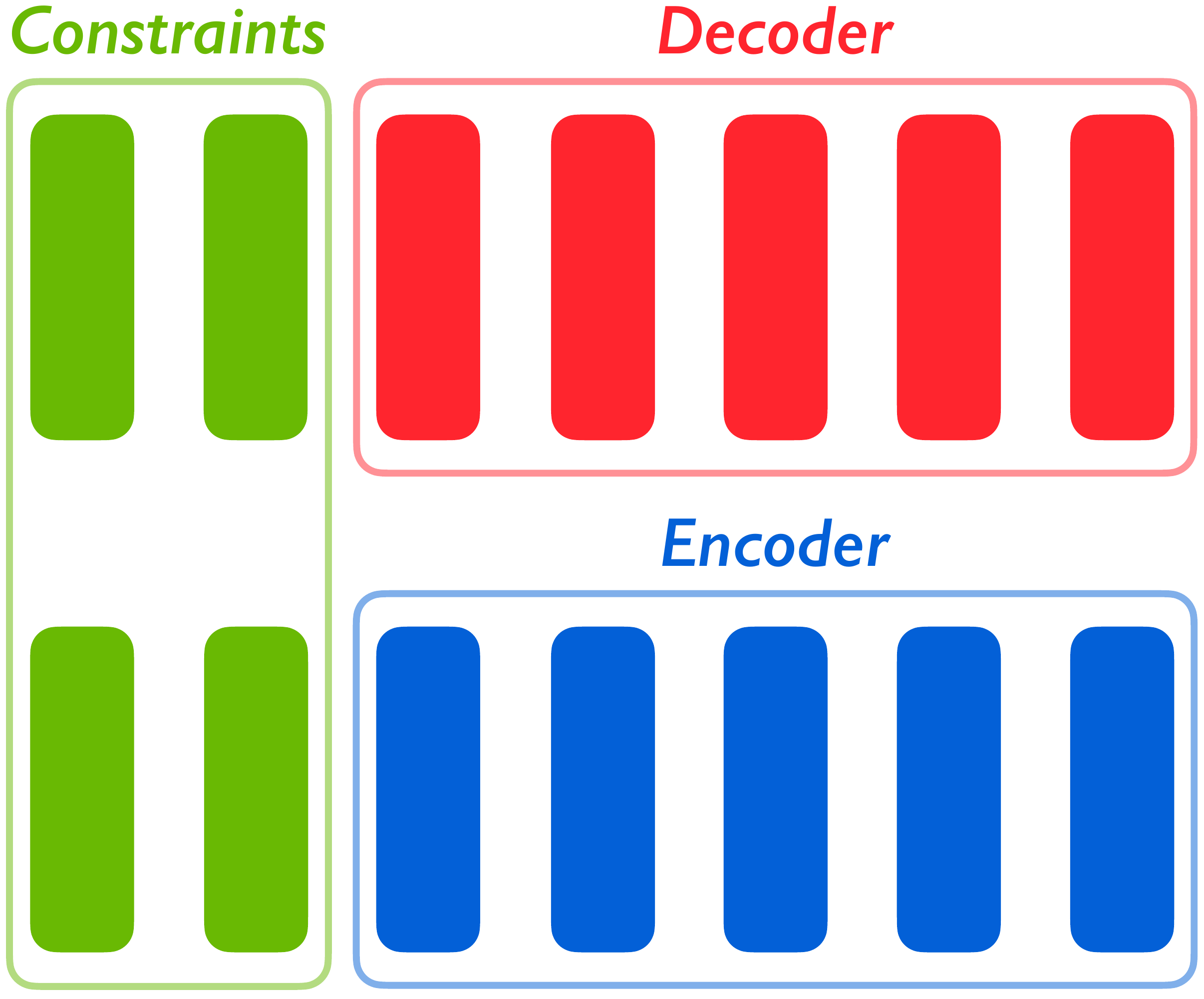}}
    \hspace{0.01\textwidth}
    \subfloat[Integration into enc. self-attn.]{
    \includegraphics[width=0.31\textwidth]{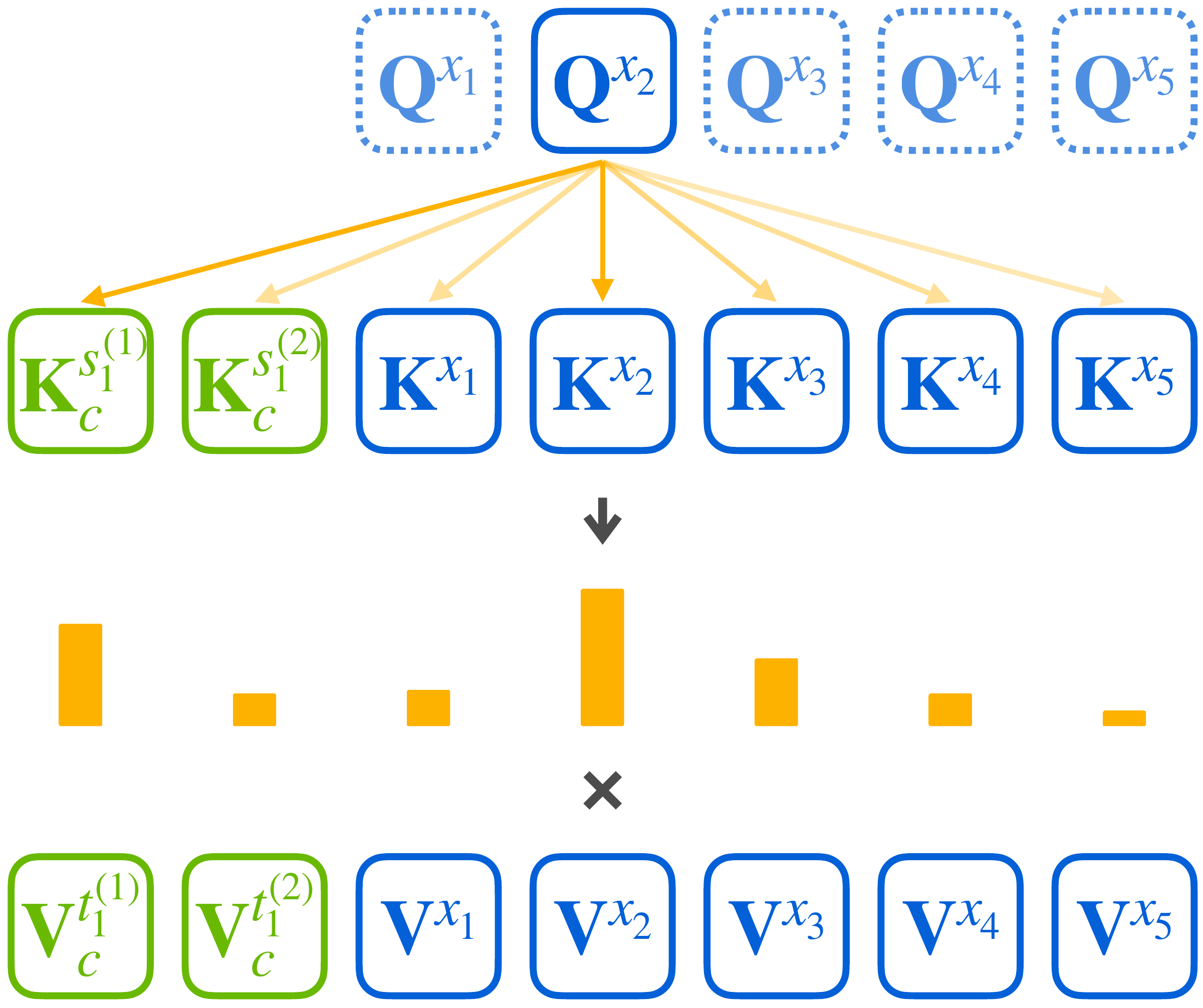}
    \label{fig:enc-attn}
    }
    \hspace{0.01\textwidth}
    \subfloat[Integration into dec. cross-attn.]{
    \includegraphics[width=0.31\textwidth]{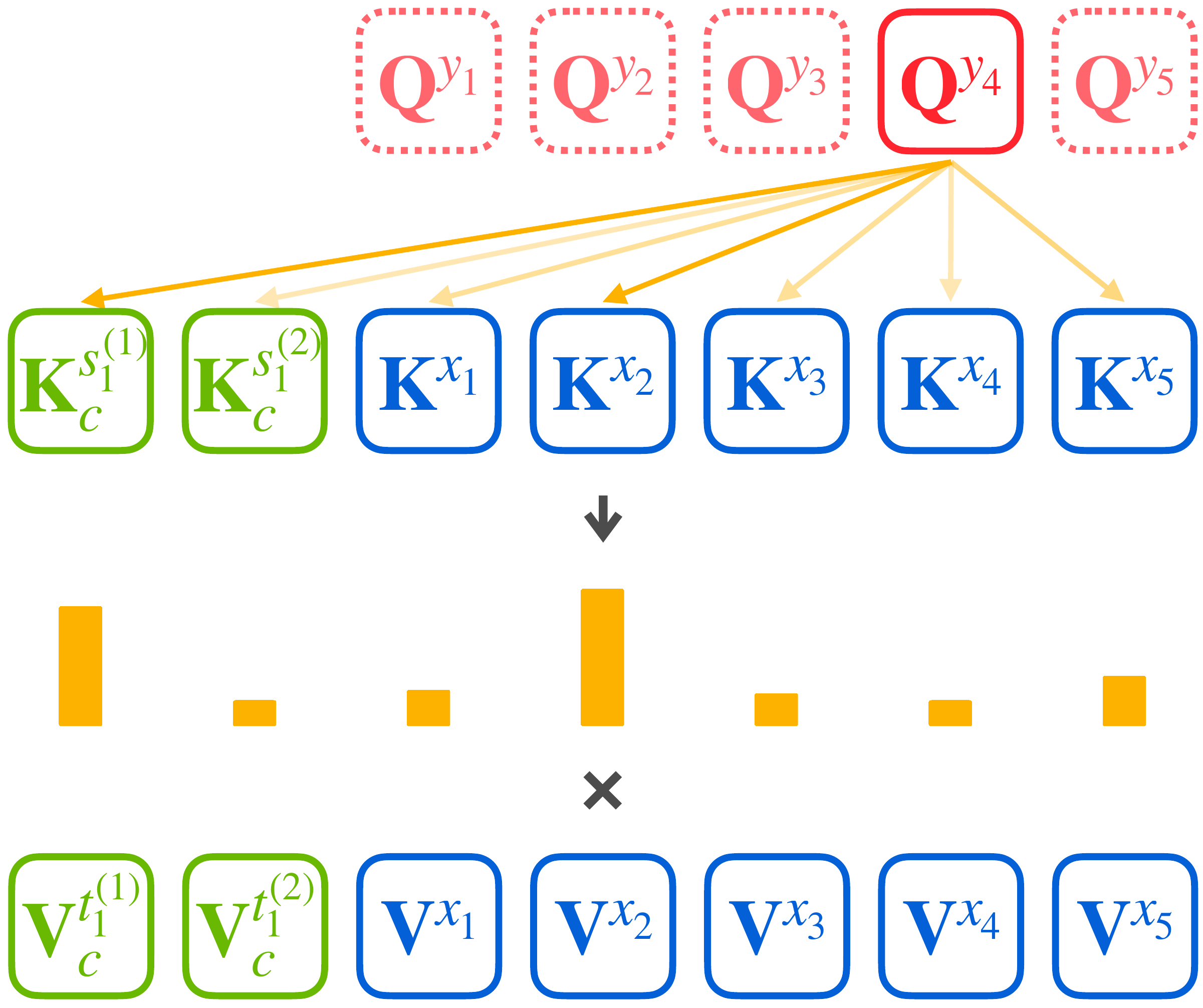}
    \label{fig:dec-attn}
    }
    \caption{Illustration of the integration of vectorized lexical constraints into both the encoder and the decoder. Blue, red, and green squares represent vectorized representations for source tokens, target tokens, and tokens of constraint pairs, respectively. The basic idea is to use source constraints as indicators to select the corresponding target constraints for each query. We only plot the attention weights for one query for simplicity.
    }
    \label{fig:example-const-attn}
\end{figure*}

\section{Approach}

This section explains how we integrate lexical constraints into NMT models. Section~\ref{sec:vec-const} illustrates the way we encode discrete constraints into continuous vectors, Section~\ref{sec:int-const} details how we integrate the vectorized constraints into NMT models, and Section~\ref{sec:loss} describes our training strategy.

\subsection{Vectorizing Lexical Constraints}
\label{sec:vec-const}

Let $\mathbf{s}=\mathbf{s}^{(1)}, \dots, \mathbf{s}^{(N)}$ be the source constraints and $\mathbf{t}=\mathbf{t}^{(1)}, \dots, \mathbf{t}^{(N)}$ be the target constraints. Given a constraint pair $\langle \mathbf{s}^{(n)}, \mathbf{t}^{(n)} \rangle$, lexically constrained translation requires that the system must translate the source phrase $\mathbf{s}^{(n)}$ into the target phrase $\mathbf{t}^{(n)}$.
Since the inner states of NMT models are all continuous vectors rather than discrete tokens, we need to vectorize the constraints before integrating them into NMT models.

For the $n$-th constraint pair $\langle \mathbf{s}^{(n)}, \mathbf{t}^{(n)} \rangle$, let $|\mathbf{s}^{(n)}|$ and $|\mathbf{t}^{(n)}|$ be the lengths of $\mathbf{s}^{(n)}$ and $\mathbf{t}^{(n)}$, respectively.
We use $\mathbf{S}^{(n)}_k \in \mathbb{R}^{d \times 1}$ to denote the vector representation of the $k$-th token in $\mathbf{s}^{(n)}$, which is the sum of word embedding and positional embedding~\cite{Vaswani:2017:Transformer}. Therefore, the matrix representation of $\mathbf{s}^{(n)}$ is given by:
\begin{equation}
    \mathbf{S}^{(n)} = \left [  \mathbf{S}^{(n)}_1 ; \dots; \mathbf{S}^{(n)}_{|\mathbf{s}^{(n)}|} \right ],
\end{equation}
where $\mathbf{S}^{(n)} \in \mathbb{R}^{d \times |\mathbf{s}^{(n)}|}$ is the concatenation of all vector representations of tokens in $\mathbf{s}^{(n)}$. Similarly, the matrix representation of the target constraint $\mathbf{t}^{(n)}$ is $\mathbf{T}^{(n)} \in \mathbb{R}^{d \times |\mathbf{t}^{(n)}|}$. Note that the positional embedding for each constraint is calculated independently, which is also independent of the positional embeddings of the source sentence $\mathbf{x}$ and the target sentence $\mathbf{y}$.

\subsection{Integrating Vectorized Constraints}
\label{sec:int-const}

We adopt Transformer~\cite{Vaswani:2017:Transformer} as our NMT model, which is nowadays one of the most popular and effective NMT models~\cite{Liu:2020:mBART}. Typically, a Transformer consists of an encoder, a decoder, and an output layer, of which the encoder and decoder map discrete tokens into vectorized representations and the output layer converts such representations into token-level probability distributions. We propose to utilize the attention modules to integrate the constraints into the encoder and decoder and use a plug-in module to integrate constraints into the output layer. We change the formal representation of our model from $P(\mathbf{y}|\mathbf{x}; \bm{\theta})$ to $P(\mathbf{y}|\mathbf{x}, \mathbf{s}, \mathbf{t}; \bm{\theta})$ to indicate that the model explicitly considers lexical constraints when estimating probability.

\paragraph{Constraint-Related Keys and Values}
We propose to map source and target constraints into additional keys and values, which are called {\em constraint-related keys and values}, in order to distinguish from the original keys and values in vanilla attention modules.
In practice, source and target constraints may have different lengths and they are usually not monotonically aligned~\cite{Du:2021:OAXE}, making it challenging to directly convert the constraints into keys and values. To fix this problem, We adopt a multi-head attention layer~\cite{Vaswani:2017:Transformer} to align the bilingual constraints. The constraint-related keys and values for the $n$-th constraint pair are given by
\begin{equation}
\begin{split}
\label{eq:kv-attn}
    \mathbf{K}_{c}^{(n)} &= \mathbf{S}^{(n)}, \\
    \mathbf{V}_{c}^{(n)} &= \mathrm{attn} \left ( \mathbf{S}^{(n)}, \mathbf{T}^{(n)}, \mathbf{T}^{(n)} \right ),
\end{split}
\end{equation}
where $\mathbf{K}_{c}^{(n)} \in \mathbb{R}^{d \times |\mathbf{s}^{(n)}|}$ and $\mathbf{V}_{c}^{(n)} \in \mathbb{R}^{d \times |\mathbf{s}^{(n)}|}$. $\mathrm{attn}(\mathbf{Q}, \mathbf{K}, \mathbf{V})$ denotes the multi-head attention function. Note that the resulting $\mathbf{K}_{c}^{(n)}$ and $\mathbf{V}_{c}^{(n)}$ are of the same shape. $\mathbf{V}_{c}^{(n)}$ can be seen as a re-distributed version of the representation of target constraints. The constraint-related keys and values of each constraint pair are calculated separately and then concatenated together:
\begin{equation}
\begin{split}
\label{eq:const-related-kv}
    \mathbf{K}_c &= [ \mathbf{K}_c^{(1)}; \dots ; \mathbf{K}_c^{(N)} ], \\
    \mathbf{V}_c &= [ \mathbf{V}_c^{(1)}; \dots ; \mathbf{V}_c^{(N)} ],
\end{split}
\end{equation}
where $\mathbf{K}_{c} \in \mathbb{R}^{d \times |\mathbf{s}|}$ and $\mathbf{V}_{c} \in \mathbb{R}^{d \times |\mathbf{s}|}$. $|\mathbf{s}|$ is the total length of all the $N$ source constraints.

\paragraph{Integration into the Encoder} The encoder of Transformer is a stack of $I$ identical layers, each layer contains a self-attention module to learn context-aware representations. For the $i$-th layer, the self-attention module can be represented as
\begin{equation}
\label{eq:enc-self-attn}
    \mathrm{attn}\left (
        \mathbf{H}^{(i-1)}_{\mathrm{enc}},
        \mathbf{H}^{(i-1)}_{\mathrm{enc}}, 
        \mathbf{H}^{(i-1)}_{\mathrm{enc}} 
    \right ),
\end{equation}
where $\mathbf{H}^{(i-1)}_{\mathrm{enc}} \in \mathbb{R}^{d \times |\mathbf{x}|}$ is the output of the $(i-1)$-th layer, and $\mathbf{H}^{(0)}_{\mathrm{enc}}$ is initialized as the sum of word embedding and positional embedding~\cite{Vaswani:2017:Transformer}. For different layers, $\mathbf{H}^{(i-1)}_{\mathrm{enc}}$ may lay in various manifolds, containing different levels of information~\cite{Voita:2019:Bottom-Up}. Therefore, we should adapt the constraint-related keys and values for each layer before the integration. We use a two-layer adaptation network to do this:
\begin{equation}
\begin{split}
\label{eq:enc-adapt}
    \mathbf{K}_{\mathrm{c4enc}}^{(i)} &= [\mathrm{adapt}(\mathbf{K}_c); \mathbf{H}^{(i-1)}_{\mathrm{enc}}], \\
    \mathbf{V}_{\mathrm{c4enc}}^{(i)} &= [\mathrm{adapt}(\mathbf{V}_c); \mathbf{H}^{(i-1)}_{\mathrm{enc}}], \\
\end{split}
\end{equation}
where $\mathrm{adapt}(\cdot)$  denotes the adaptation network, which consists of two linear transformations with shape $d \times d$ and a ReLU activation in between. The adaptation networks across all layers are independent of each other.
$\mathbf{K}_{\mathrm{c4enc}}^{(i)} \in \mathbb{R}^{d \times (|\mathbf{s}| + |\mathbf{x}|)}$ and $\mathbf{V}_{\mathrm{c4enc}}^{(i)} \in \mathbb{R}^{d \times (|\mathbf{s}| + |\mathbf{x}|)}$ are the constraint-aware keys and values for the $i$-th encoder layer, respectively. The vanilla self-attention module illustrated in Eq.~(\ref{eq:enc-self-attn}) is revised into the following form:
\begin{equation}
    \mathrm{attn}\left (
        \mathbf{H}^{(i-1)}_{\mathrm{enc}},
        \mathbf{K}_{\mathrm{c4enc}}^{(i)},
        \mathbf{V}_{\mathrm{c4enc}}^{(i)} 
    \right ).
\end{equation}

Figure~\ref{fig:enc-attn} plots an example of the integration into the encoder self-attention.

\begin{figure}[ht]
    \centering
    \includegraphics[width=0.42\textwidth]{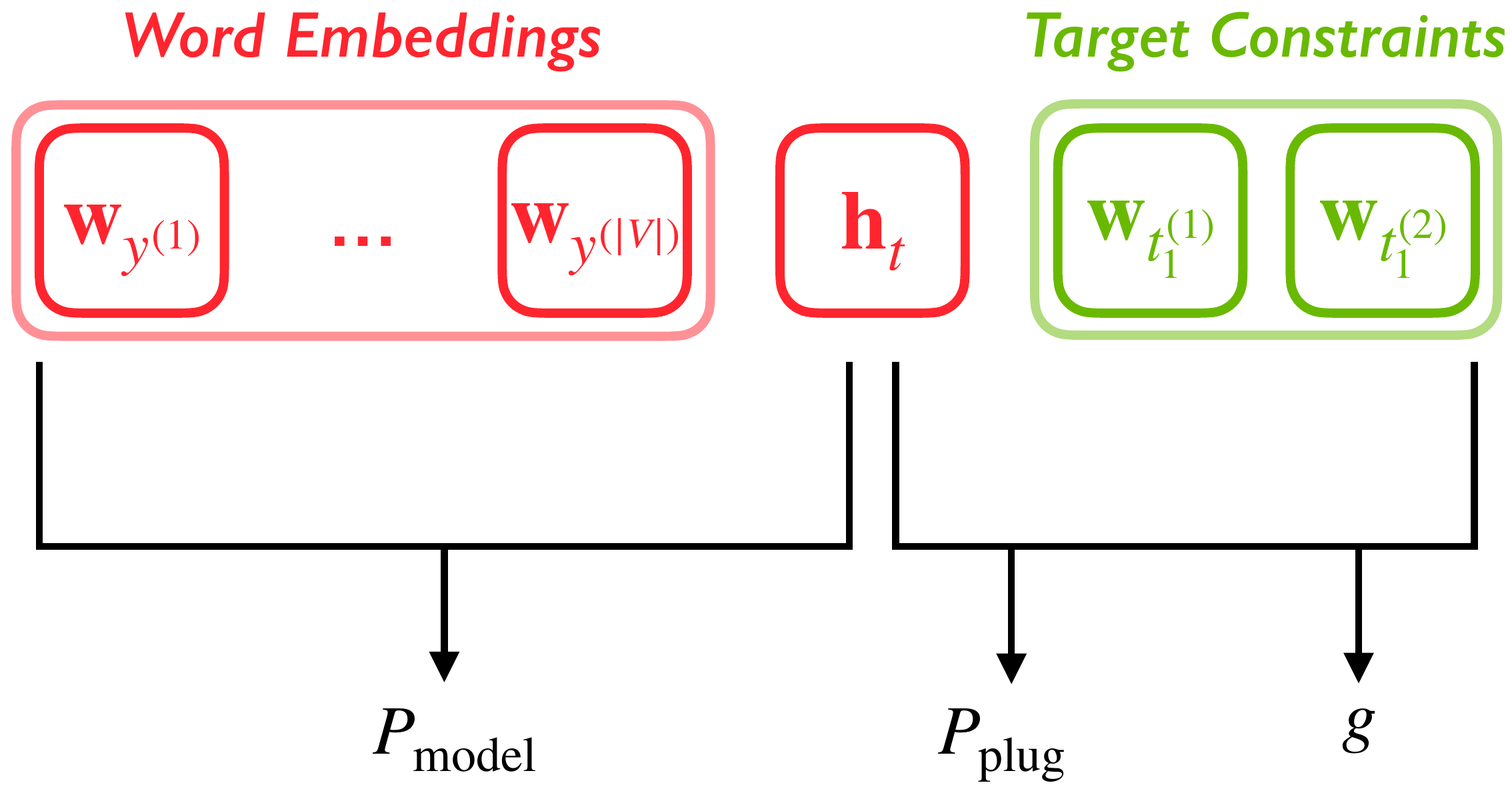}
    \caption{Illustration of the integration into the output layer. Please refer to Eq (\ref{eq:p-model}), (\ref{eq:p-plug}), and (\ref{eq:gate}) for the definition of $P_{\mathrm{model}}$, $P_{\mathrm{plug}}$, and $g$, respectively.}
    \label{fig:output-plug}
\end{figure}

\paragraph{Integration into the Decoder} The integration into the decoder is similar to that into the encoder, the major difference is that we use the cross-attention module to model constraints for the decoder. 
The decoder of the Transformer is a stack of $J$ identical layers, each of which is composed of a self-attention, a cross-attention, and a feed-forward module. We integrate vectorized constraints into the cross-attention module for the decoder. Formally, the vanilla cross-attention is given by
\begin{equation}
\label{eq:dec-cross-attn}
    \mathrm{attn}\left (
        \mathbf{S}^{(j)}_{\mathrm{dec}},
        \mathbf{H}^{(I)}_{\mathrm{enc}}, 
        \mathbf{H}^{(I)}_{\mathrm{enc}}
    \right ),
\end{equation}
where $\mathbf{S}^{(j)}_{\mathrm{dec}} \in \mathbb{R}^{d \times |\mathbf{y}|}$ is the output of the self-attention module in the $j$-th decoder layer, and $\mathbf{H}^{(I)}_{\mathrm{enc}} \in \mathbb{R}^{d \times |\mathbf{x}|}$ is the output of the last encoder layer. We adapt the constraint-related keys and values to match the manifold in the $j$-th decoder layer:
\begin{equation}
\begin{split}
\label{eq:dec-adapt}
    & \mathbf{K}_{\mathrm{c4dec}}^{(j)} = [\mathrm{adapt}(\mathbf{K}_c); \mathbf{H}^{(I)}_{\mathrm{enc}}], \\
    & \mathbf{V}_{\mathrm{c4dec}}^{(j)} = [\mathrm{adapt}(\mathbf{V}_c); \mathbf{H}^{(I)}_{\mathrm{enc}}]. \\
\end{split}
\end{equation}

Then we revise the vanilla cross-attention (Eq.~(\ref{eq:dec-cross-attn})) into the following form:
\begin{equation}
    \mathrm{attn}\left (
        \mathbf{S}^{(j)}_{\mathrm{dec}},
        \mathbf{K}_{\mathrm{c4dec}}^{(j)},
        \mathbf{V}_{\mathrm{c4dec}}^{(j)}
    \right ).
\end{equation}

Figure~\ref{fig:dec-attn} plots an example of the integration into the decoder cross-attention. 

\paragraph{Integration into the Output Layer} In vanilla Transformer, an output layer is employed to convert the output of the last decoder layer into token-level probabilities. Let $\mathbf{h}_t \in \mathbb{R}^{d \times 1}$ be the decoder output at the $t$-th time step, the output probability of the Transformer model is defined as
\begin{equation}
    \label{eq:p-model}
    P_{\mathrm{model}}(y|\mathbf{y}_{<t}, \mathbf{x}, \mathbf{s}, \mathbf{t}; \bm{\theta}) = \mathrm{softmax}
    \left ( \mathbf{h}_{t}^{\top} \mathbf{W} \right ),
\end{equation}
where $\mathbf{W} \in \mathbb{R}^{d \times |\mathcal{V}|}$ is the output embedding matrix and $|\mathcal{V}|$ is the vocabulary size.
Inspired by the plug-and-play method~\cite{Pascual:2021:PlugAndPlay} in the field of controlled text generation~\cite{Dathathri:2020:Plug,Pascual:2021:PlugAndPlay}, we introduce an additional probability distribution over the vocabulary to better generate constrained tokens:
\begin{equation}
\label{eq:p-plug}
\begin{split}
    & P_{\mathrm{plug}}(y|\mathbf{y}_{<t}, \mathbf{x}, \mathbf{s}, \mathbf{t}; \bm{\theta})\\
    &~~= \left \{
    \begin{split}
        0 \quad \quad \quad \quad \quad \quad \quad \quad \quad \quad \quad & \quad y \notin \mathbf{t} \\
        \mathrm{max} \left(0, \mathrm{cos}\left(\frac{\mathbf{w}_{y}}{|\mathbf{w}_{y}|}, \frac{\mathbf{h}_{t}}{|\mathbf{h}_{t}|}\right)\right) & \quad y \in \mathbf{t} \\
    \end{split}
    \right.
\end{split},
\end{equation}
where $\mathbf{w}_{y} \in \mathbf{R}^{d \times 1}$ is the word embedding of token $y$ and $\mathbf{t}$ is the sequence of all the target-side constrained tokens. We also use a gating sub-layer to control the strength of the additional probability:
\begin{equation}
\label{eq:gate}
\begin{split}
    &  g(y, \mathbf{h}_{t}) \\
    & ~~ = \mathrm{sigmoid} \left(
        \mathrm{tanh} \left( \left[
            \mathbf{w}_{y}^{\top} \mathbf{W}_1; \mathbf{h}_t^{\top} \mathbf{W}_2
        \right]\right) \mathbf{W}_3
    \right),
\end{split}
\end{equation}
where $\mathbf{W}_1 \in \mathbf{R}^{d \times d}$, $\mathbf{W}_2 \in \mathbf{R}^{d \times d}$, and $\mathbf{W}_3 \in \mathbf{R}^{2d \times 1}$ are three trainable linear transformations. The final output probability is given by
\begin{equation}
\begin{split}
    & P(y|\mathbf{y}_{<t}, \mathbf{x}, \mathbf{s}, \mathbf{t}; \bm{\theta}) \\
    & ~~ = \left(1 - g(y, \mathbf{h}_{t})\right) P_{\mathrm{model}}(y|\mathbf{y}_{<t}, \mathbf{x}, \mathbf{s}, \mathbf{t}; \bm{\theta}) \\
    & ~~ + g(y, \mathbf{h}_{t}) P_{\mathrm{plug}}(y|\mathbf{y}_{<t}, \mathbf{x}, \mathbf{s}, \mathbf{t}; \bm{\theta}).
\end{split}
\end{equation}

\subsection{Training and Inference}
\label{sec:loss}
\paragraph{Training} The proposed constraint-aware NMT model should not only generate pre-specified constraints but also maintain or improve the translation quality compared with vanilla NMT models. We thus propose to distinguish between constraint tokens and constraint-unrelated tokens during training. Formally, the training objective is given by 
\begin{equation}
\begin{split}
    & L(\mathbf{y}|\mathbf{x}, \mathbf{s}, \mathbf{t}; \bm{\theta}) \\
    & ~~ = \alpha \sum_{y_t \in \mathbf{y} \cap \mathbf{t}} \log P(y_t|\mathbf{y}_{<t}, \mathbf{x}, \mathbf{s}, \mathbf{t}; \bm{\theta}) \\
    & ~~ + ~ \beta \sum_{y_t \in \mathbf{y} \setminus \mathbf{t}} \log P(y_t|\mathbf{y}_{<t}, \mathbf{x}, \mathbf{s}, \mathbf{t}; \bm{\theta}),
\end{split}
\end{equation}
where $\alpha$ and $\beta$ are hyperparameters to balance the learning of constraint generation and translation.

We can divide the parameter set of the whole model into two subsets: $\bm{\theta} = \bm{\theta}_{v} \cup \bm{\theta}_{c}$, where $\bm{\theta}_{v}$ is a set of original vanilla model parameters and $\bm{\theta}_c$ is a set of newly-introduced parameters that are used to vectorize and integrate lexical constraints.\footnote{$\bm{\theta}_{c}$ includes parameters of the attention presented in Eq~(\ref{eq:kv-attn}), the adaptation networks described in Eq~(\ref{eq:enc-adapt}) and (\ref{eq:dec-adapt}), and the gating sub-layer illustrated in Eq~(\ref{eq:gate}).} Since $\bm{\theta}_c$ is significantly smaller than $\bm{\theta}_{v}$, it requires much less training iterations. Therefore, we adopt the strategy of two-stage training~\cite{Tu:2018:Cache,Zhang:2018:Document} for model optimization. Specifically, we optimize $\bm{\theta}_v$ using the standard NMT training objective~\cite{Bahdanau:2015:RNNSearch,Vaswani:2017:Transformer} at the first stage and then learn the whole model $\bm{\theta}$ at the second stage. The second stage is significantly shorter than the first stage, we will give more details in Section~\ref{sec:exp-setup}.

\paragraph{Inference} As discussed in Section~\ref{sec:nmt}, the inference process is composed of two sub-processes: probability estimation and candidate selection. In this work, we aim to improve the probability estimation sub-process and our method is orthogonal to constrained decoding algorithms~\cite{Hokamp:2017:Lexical,Post:2018:Lexical,Hu:2019:Lexical}, which instead focus on candidate selection. Therefore, we can employ not only beam search but also constrained decoding algorithms at inference time. We use VDBA~\cite{Hu:2019:Lexical} as the default constrained decoding algorithm, which supports batched inputs and is significantly faster than most other counterparts~\cite{Hokamp:2017:Lexical,Post:2018:Lexical,Hasler:2018:Lexical}.

\section{Experiments}

\subsection{Setup}
\label{sec:exp-setup}
\paragraph{Training Data} In this work, we conduct experiments on Chinese$\Leftrightarrow$English (Zh$\Leftrightarrow$En) and German$\Leftrightarrow$English (De$\Leftrightarrow$En) translation tasks. For Zh$\Leftrightarrow$En, the training set contains 1.25M sentence pairs from LDC\footnote{The total training set for Zh$\Leftrightarrow$En is composed of LDC2002E18, LDC2003E07, LDC2003E14, part of LDC2004T07, LDC2004T08 and LDC2005T06.}. For De$\Leftrightarrow$En, the training set is from the WMT 2014 German$\Leftrightarrow$English translation task, which consists of 4.47M sentence pairs. We apply BPE~\cite{Sennrich:2016:BPE} with 32K joint merge operations for both Zh$\Leftrightarrow$En and De$\Leftrightarrow$En.

\paragraph{Evaluation Data} Following \citet{Chen:2021:Lexical}, we evaluate our approach on the test sets with human-annotated alignments, which are widely used in related studies~\cite{Chen:2020:Align,Chen:2021:Align}.
We find the alignment test sets have significant overlaps with the corresponding training sets, which is not explicitly stated in previous works. In this work, we remove the training examples that are covered by the alignment test sets.
For Zh$\Leftrightarrow$En, we use the alignment datasets from \citet{Liu:2005:Align}\footnote{\url{http://nlp.csai.tsinghua.edu.cn/~ly/systems/TsinghuaAligner/TsinghuaAligner.html}}, 
in which the validation and test sets both contain 450 sentence pairs. For De$\Leftrightarrow$En, we use the alignment dataset from \citet{Zenkel:2020:Align}\footnote{\url{https://github.com/lilt/alignment-scripts}}
as the test set, which consists of 508 sentence pairs. Since there is no human-annotated alignment validation sets for De$\Leftrightarrow$En, we use \verb|fast-align|\footnote{\url{https://github.com/clab/fast_align}} to annotate the newstest 2013 as the validation set for De$\Leftrightarrow$En. 

\paragraph{Lexical Constraints} In real-world applications, lexical constraints are usually provided by human translators. We follow \citet{Chen:2021:Lexical} to simulate the practical scenario by sampling constraints from the phrase pairs that are extracted from parallel data using alignments. The script for phrase pair extraction is publicly available.\footnote{\url{https://github.com/ghchen18/cdalign/blob/main/scripts/extract_phrase.py}} For the validation and test sets of Zh$\Leftrightarrow$En and the test set of De$\Leftrightarrow$En, we use human-annotated alignments to extract phrase pairs. For the training corpora in both Zh$\Leftrightarrow$En and De$\Leftrightarrow$En, we use \verb|fast-align| to firstly learn an alignment model and then use the model to automatically annotate the alignments. The validation set of De$\Leftrightarrow$En is also annotated by the alignment model learned on the corresponding training corpus. We use the same strategy as \citet{Chen:2021:Lexical} to sample constraints from the extracted phrase pairs. More concretely, the number of constraints in each sentence is up to 3. The length of each constrained phrase is uniformly sampled among 1 and 3. For each sentence pair, all the constraint pairs are shuffled and then supplied to the model in an unordered manner.

\begin{table*}[t]
    \centering
    \scalebox{1.0}{
    \begin{tabular}{l lllll rrrrr}
    \toprule
    \multirow{2}{*}{\bf Method} & \multicolumn{5}{c}{\bf BLEU} & \multicolumn{5}{c}{\bf CSR (\%)} \\
    \cmidrule(lr){2-6} \cmidrule(lr){7-11}
    & \em Z$\rightarrow$E & \em E$\rightarrow$Z & \em D$\rightarrow$E & \em E$\rightarrow$D & \em Avg. & \em Z$\rightarrow$E & \em E$\rightarrow$Z & \em D$\rightarrow$E & \em E$\rightarrow$D & \em Avg. \\
    \midrule
    Vanilla & 30.4 & 56.1 & 31.7 & 24.5 & 35.7 & 26.9  & 30.5  & 19.2  & 13.8  & 22.6  \\
    \cmidrule(lr){1-1} \cmidrule(lr){2-6} \cmidrule(lr){7-11}
    VDBA & 31.6 & 56.4 & 35.0 & 27.9 & 37.7 & \em 99.4  & \em 98.9  & \em 100.0  & \em 100.0  & \em 99.6  \\
    Replace & 33.6 & 58.3 & 34.9 & 28.1 & 38.7 & 89.7 & 90.2  & 93.2  & 90.7  & 91.0  \\
    CDAlign & 32.1 & 58.0 & 35.2 & 28.3$^\star$ & 38.4 & 87.4  & 90.0  & 96.8  & 95.7  & 92.5  \\
    \cmidrule(lr){1-1} \cmidrule(lr){2-6} \cmidrule(lr){7-11}
    Ours & \bf 34.4 & \bf 59.1 & \bf 35.9 & \bf 28.8 & \bf 39.6 & \em 99.4  & \em 98.9  & \em 100.0  & \em 100.0  & \em 99.6  \\
    \bottomrule
    \end{tabular}}
    \caption{Results on lexically constrained test sets. "{\em Z}$\rightarrow${\em E}" denotes Zh$\rightarrow$En, and "{\em D}$\rightarrow${\em E}" denotes De$\rightarrow$En. The {\bf best BLEU} in each column is highlighted in bold, and "$\star$" indicates no significant difference with the method achieving the best BLEU. The {\em best CSR} in each column is italicized.}
    \label{tab:main-res-lexical}
\end{table*}

\paragraph{Model Configuration} We use the base setting~\cite{Vaswani:2017:Transformer} for our model. Specifically, the hidden size $d$ is 512 and the depths of both the encoder and the decoder are 6. Each multi-head attention module has 8 individual attention heads. Since our method introduces additional parameters, we use a larger model with an 8-layer encoder and an 8-layer decoder to assimilate the parameter count for the baselines. For Zh$\Leftrightarrow$En, we optimize $\bm{\theta}_v$ for 50K iterations at the first stage and then optimize $\bm{\theta}$ for 10K iterations at the second stage. For a fair comparison, we train the baselines for 60K iterations in total. For De$\Leftrightarrow$En, we optimize $\bm{\theta}_v$ for 90K iterations then optimize $\bm{\theta}$ for 10K iterations. The baselines are trained for 100K iterations.
All the involved models are optimized by Adam~\cite{Kingma:2015:CoRR}, with $\beta_{1}=0.9$, $\beta_{2}=0.98$ and $\epsilon=10^{-9}$. The dropout rate is set to 0.3 for Zh$\Leftrightarrow$En and 0.1 for De$\Leftrightarrow$En. Label smoothing is employed and the smoothing penalty is set to 0.1 for all language pairs.
We use the same learning rate schedule as \citet{Vaswani:2017:Transformer}. 
All models are trained on 4 NVIDIA V100 GPUs and evaluated on 1 NVIDIA V100 GPU. During training, each mini batch contains roughly 32K tokens in total across all GPUs.
We set the values of $\alpha$ and $\beta$ based on the results on the validation set. Specifically, for models using VDBA, we set $\alpha=\beta=0.5$, while for models using beam search, we set $\alpha=0.8$ and $\beta=0.2$.  The beam size is set to $4$ during inference.

\paragraph{Baselines} We compare our approach with three representative baselines:
\begin{itemize}
    \item {\em VDBA}~\cite{Hu:2019:Lexical}: dynamically devoting part of the beam for constraint-related hypotheses at inference time;
    \item {\em Replace}~\cite{Song:2019:Lexical}: directly replacing source constraints in the training data with their corresponding target constraints. The model is also improved with pointer network;
    \item {\em CDAlign}~\cite{Chen:2021:Lexical}: explicitly using an alignment model to decide the position to insert target constraints during inference.
\end{itemize}

\paragraph{Evaluation Metrics} We evaluate the involved methods using the following two metrics:
\begin{itemize}
    \item {\em BLEU}: we use sacreBLEU\footnote{Signature for Zh$\rightarrow$En, De$\rightarrow$En, and En$\rightarrow$De: nrefs:1 | case:mixed | eff:no | tok:13a | smooth:exp | version:2.0.0. Signature for En$\rightarrow$Zh: nrefs:1 | case:mixed | eff:no | tok:zh | smooth:exp | version:2.0.0.}~\cite{Post:2018:BLEU} to report the BLEU score;
    \item {\em Copying Success Rate (CSR)}: We follow \citet{Chen:2021:Lexical} to use the percentage of constraints that are successfully generated in the translation as the CSR, which is calculated at word level after removing the BPE separator.
\end{itemize}
We use \verb|compare-mt|~\cite{Neubig:2019:Compare} for significance testing, with $\mathrm{bootstrap}=1000$ and $\mathrm{prob\_thresh}=0.05$.

\subsection{Main Results}

Table~\ref{tab:main-res-lexical} shows the results of lexically constrained translation on test sets of all four translation tasks. All the investigated methods can effectively improve the CSR over the vanilla Transformer. The CSR of VDBA on Zh$\Leftrightarrow$En is not 100.0\% for the reason that some target constraints contain out-of-vocabulary tokens.
Replace~\cite{Song:2019:Lexical} achieves better BLEU scores on three translation directions (i.e., Zh$\Leftrightarrow$En and En$\rightarrow$De) than VDBA, but its CSR is much lower. CDAlign~\cite{Chen:2021:Lexical} also performs better than Replace on average regarding CSR. Our method consistently outperforms all the three baselines across the four translation directions in terms of BLEU, demonstrating the necessity of integrating vectorized constraints into NMT models. 
Decoding with VDBA, we also achieve the highest CSR.
To disentangle the effect of integrating vectorized constraints and VDBA, we also report the result of our model using beam search in Table~\ref{tab:beam-search}. Decoding with beam search, our model can also achieve a better BLEU score than the baselines and the CSR is higher than both Replace and CDAlign on average.

\begin{table}[t]
    \centering
    \scalebox{0.93}{
    \begin{tabular}{l rrrrr}
    \toprule
    \bf Metric & \em Z$\rightarrow$E & \em E$\rightarrow$Z & \em D$\rightarrow$E & \em E$\rightarrow$D & \em Avg. \\
    \midrule
   \bf BLEU & 34.5 & 59.5 & 35.7 & 28.6 & 39.6 \\
    \bf CSR (\%) & 94.6 & 92.4 & 97.3 & 93.3 & 94.4 \\
    \bottomrule
    \end{tabular}}
    \caption{Performance on lexically constrained test sets of the proposed model decoding with beam search.}
    \label{tab:beam-search}
\end{table}

\begin{table}[t]
    \centering
    \scalebox{0.93}{
    \begin{tabular}{c c c c c}
    \toprule
    \multicolumn{2}{c}{\bf Integration} & \multirow{2}{*}{\bf DA} & \multirow{2}{*}{\bf BLEU} & \multirow{2}{*}{\bf CSR (\%)} \\
    \cmidrule(lr){1-2}
    \em Attention & \em Output \\
    \midrule
    \checkmark & \texttimes & \multirow{3}{*}{B} & 34.3 & 91.5 \\
    \texttimes & \checkmark & & 32.6 & 61.9 \\
    \checkmark & \checkmark & & \bf 34.5 & \em 94.2 \\
    \cmidrule(lr){1-3} \cmidrule(lr){4-5}
    \checkmark & \texttimes & \multirow{3}{*}{V} & 34.4 & \em 98.6 \\
    \texttimes & \checkmark & & 33.5 & \em 98.6 \\
    \checkmark & \checkmark & & \bf 34.7 & \em 98.6 \\
    \bottomrule
    \end{tabular}}
    \caption{Effect of different components. The results are reported on the Zh$\rightarrow$En validation set. "{\em Attention}": the constraint integration of attention modules. "{\em Output}": the integration into the output layer. "DA": decoding algorithm. "B": beam search. "V": VDBA.}
    \label{tab:ablation}
\end{table}

\subsection{Ablation Study}

We investigate the effect of different components through an ablation study, the results are shown in Table~\ref{tab:ablation}. We find that only integrating lexical constraints into attention can significantly improve the CSR over the vanilla model (91.5\% vs. 25.5\%), which is consistent with our motivation that the correspondence between keys and values is naturally suitable for modeling the relation between source and target constraints. Plugging target constraints into the output layer can further improve the performance, but the output plug-in itself can only generate 61.9\% of constraints. When decoding with VDBA, combining both the two types of integration 
achieves the best BLEU score, indicating that every component is important for the model to translate with constraints.

\begin{table}[ht]
    \centering
    \subfloat[BLEU score on code-switched test sets.]{
        \scalebox{0.93}{
        \begin{tabular}{l llll}
        \toprule
        \multirow{2}{*}{\bf Method} & \multicolumn{4}{c}{\bf BLEU} \\
        \cmidrule(lr){2-5}
        & \em Z$\rightarrow$E & \em E$\rightarrow$Z & \em D$\rightarrow$E & \em E$\rightarrow$D \\
        \midrule
        Vanilla & 29.7 & 52.9 & 29.3 & 23.5 \\
        \cmidrule(lr){1-1} \cmidrule(lr){2-5}
        VDBA & 31.4 & 54.0 & 31.8 & 26.2 \\
        Replace & 31.8 & 56.0 & 31.1 & 25.7 \\
        CDAlign & 32.3 & 55.1 & 32.2 & 26.4 \\
        \cmidrule(lr){1-1} \cmidrule(lr){2-5}
        Ours & 33.1$^\star$ & 56.5$^\star$ & 32.7$^\star$ & 27.0$^\star$ \\
        ~~w/o VDBA & \bf 33.5 & \bf 56.7 & \bf 32.9 & \bf 27.1 \\
        \bottomrule
        \end{tabular}}
    }
    
    \subfloat[CSR on code-switched test sets.]{
        \scalebox{0.93}{
        \begin{tabular}{l rrrr}
        \toprule
        \multirow{2}{*}{\bf Method} & \multicolumn{4}{c}{\bf CSR (\%)} \\
        \cmidrule(lr){2-5}
        & \em Z$\rightarrow$E & \em E$\rightarrow$Z & \em D$\rightarrow$E & \em E$\rightarrow$D \\
        \midrule
        Vanilla & 12.7 & 14.4 & 10.2 & 8.7 \\
        \cmidrule(lr){1-1} \cmidrule(lr){2-5}
        VDBA & \em 99.4 & \em 98.5 & \em 100.0 & \em 100.0 \\
        Replace & 44.8 & 47.4 & 44.9 & 44.1 \\
        CDAlign & 89.3 & 89.5 & 95.2 & 91.1 \\
        \cmidrule(lr){1-1} \cmidrule(lr){2-5}
        Ours & \em 99.4 & \em 98.5 & \em 100.0 & \em 100.0 \\
        ~~w/o VDBA & 96.1 & 94.6 & 97.6 & 95.8 \\
        \bottomrule
        \end{tabular}}
    }
    \caption{Results on the code-switched translation task.}
    \label{tab:code-switch}
    \end{table}

\subsection{Code-Switched Translation}

\paragraph{Task Description and Data Preparation} An interesting application of lexically constrained machine translation is code-switched translation, of which the output contains terms across different languages. Figure~\ref{fig:prompt-example} shows an example of code-switched translation, where the output Chinese sentence should include the English token "Beatles". Code-switched machine translation is important in many scenarios, such as entity translation~\cite{Li:2018:Lexical} and the translation of sentences containing product prices or web URLs~\cite{Chen:2021:Lexical}. In this work, we evaluate the performance of several approaches on code-switched machine translation. The parallel data and extracted constraint pairs for each language pair are the same as those used in the lexically constrained translation task. To construct the training and evaluation data for code-switched translation, we randomly replace 50\% of the target constraints with their corresponding source constraints. The target sentence is also switched if it contains switched target constraints. 

\paragraph{Results} Table~\ref{tab:code-switch} gives the results of the code-switched translation task. The CSR of Replace~\cite{Song:2019:Lexical} is lower than 50\% across all the four translation directions, indicating that simply replacing the training data can not handle the code-switched translation. A potential reason is that it is difficult for the NMT model to decide whether to translate or copy some source phrases in the input sentence. Surprisingly, VDBA, CDAlign, and our method all perform well in this scenario, and our method outperforms the two baselines. These results suggest the capability of our method to cope with flexible types of lexical constraints.

\section{Discussion}

\begin{table}[t]
    \centering
    \begin{tabular}{l r r}
    \toprule
    \multirow{2}{*}{\bf Method} & \multicolumn{2}{c}{\bf Batch Size (\# Sent.)} \\
    \cmidrule(lr){2-3}
    & 1~~~ & 128~~~ \\
    \midrule
    Vanilla & 1.0\texttimes & 43.2\texttimes \\
    \cmidrule(lr){1-1} \cmidrule(lr){2-3}
    VDBA & 0.5\texttimes & 2.1\texttimes \\
    Replace & 0.9\texttimes & 40.5\texttimes \\
    CDAlign & 0.7\texttimes & n/a~~~ \\
    \cmidrule(lr){1-1} \cmidrule(lr){2-3}
    Ours & 0.5\texttimes & 2.3\texttimes \\
    ~~ w/o VDBA & 0.9\texttimes & 39.2\texttimes\\
    \bottomrule
    \end{tabular}
    \caption{Inference speed with different batch sizes.}
    \label{tab:speed}
\end{table}

\subsection{Inference Speed} We report the inference speed of each involved approach in Table~\ref{tab:speed}. The speed of Replace is close to that of the vanilla Transformer, but its CSR is much lower than other methods. Since the open-sourced implementation of CDAlign\footnote{\url{https://github.com/ghchen18/cdalign}} does not support batched decoding, we compare our method with CDAlign with $\mathrm{batch\_size}=1$. The speed of our method using beam search is faster than that of CDAlign (0.9\texttimes vs. 0.7\texttimes). When provided with batched inputs, our method can slightly speed up VDBA (2.3\texttimes vs. 2.1\texttimes). A potential reason is that the probability estimated by our model is more closely related to the correctness of the candidates, making target constraints easier to find. 

\begin{table}[t]
    \centering
    \begin{tabular}{l c c c r}
    \toprule
    \multirow{2}{*}{\bf Model} & \multirow{2}{*}{\bf DA} & \multicolumn{2}{c}{\bf Prob.} & \multirow{2}{*}{\bf ECE$^{(\downarrow)}$} \\
    \cmidrule(lr){3-4}
    & & \em All & \em Const. & \\
    \midrule
    Vanilla & \multirow{2}{*}{B} & 0.69 & 0.74 & 8.03 \\
    Ours & & 0.70 & 0.80 & \bf 7.19 \\
    \cmidrule(lr){1-2} \cmidrule(lr){3-4} \cmidrule(lr){5-5}
    Vanilla & \multirow{2}{*}{V} & 0.64 & 0.42 & 10.73 \\
    Ours & & 0.67 & 0.61 & \bf 7.72 \\
    \bottomrule
    \end{tabular}
    \caption{Inference ECE on the Zh$\rightarrow$En validation set. "{\em All}": the average probability of all predicted tokens. "{\em Const.}": the average probability of constrained tokens.}
    \label{tab:calibration}
\end{table}

\subsection{Calibration}
To validate whether the probability of our model is more accurate than vanilla models, we follow \citet{Wang:2020:InfECE} to investigate the gap between the probability and the correctness of model outputs, which is measured by the inference expected calibration error (ECE).
As shown in Table~\ref{tab:calibration}, the inference ECE of our method is much lower than that of the vanilla model, indicating that the probability of our model is more accurate than vanilla models.
To better understand the calibration of our model and the baseline model, we also estimate the average probability of all the predicted tokens and the constrained tokens. The results show that our model assigns higher probabilities to constrained tokens, which are already known to be correct.

\subsection{Memory vs. Extrapolation}

\begin{table}[t]
    \centering
    \begin{tabular}{c c c c c}
    \toprule
    \em Z$\rightarrow$E & \em E$\rightarrow$Z & \em D$\rightarrow$E & \em E$\rightarrow$D & \em Avg. \\
    \cmidrule(lr){1-5}
    38.9\% & 43.4\% & 27.7\% & 32.2\% & 35.6\% \\
    \bottomrule
    \end{tabular}
    \caption{Overlap ratio of lexical constraints between training and test sets across the four directions.}
    \label{tab:extra}
\end{table}

To address the concern that the proposed model may only memorize the constraints seen in the training set, we calculate the overlap ratio of constraints between training and test sets. As shown in Table~\ref{tab:extra}, we find that only 35.6\% of the test constraints are seen in the training data, while the CSR of our model decoding with beam search is 94.4\%. The results indicate that our method extrapolates well to constraints unseen during training.

\subsection{Case Study} Table~\ref{tab:case-study} shows some example translations of different methods. We find Replace tends to omit some constraints. Although VDBA and CDAlign can successfully generate constrained tokens, the translation quality of the two methods is not satisfying. Our result not only contains constrained tokens but also maintains the translation quality compared with the unconstrained model, confirming the necessity of integrating vectorized constraints into NMT models.

\begin{table*}[ht]
\centering
\begin{tabular}{l m{11.5cm}}
\toprule
\bf Constraints & {\color{sblue} \bf Zielsetzung} $\rightarrow$ {\color{sblue} \bf objectives},  {\color{sred} \bf Fiorella} $\rightarrow$ {\color{sred} \bf Fiorella} \\
\cmidrule(lr){1-1} \cmidrule(lr){2-2}
\bf Source & Mit der {\color{sblue} \bf Zielsetzung} des Berichtes von {\color{sred} \bf Fiorella} Ghilardotti allerdings sind wir einverstanden .  \\
\bf Reference & Even so , we do agree with the {\color{sblue} \bf objectives} of {\color{sred} \bf Fiorella} Ghilardotti's report . \\
\cmidrule(lr){1-1} \cmidrule(lr){2-2}
Vanilla & However , we agree with the aims of the Ghilardotti report . \\
\cmidrule(lr){1-1} \cmidrule(lr){2-2}
VDBA & {\color{sred} \bf Fiorella}'s Ghilardotti report , however , has our {\color{sblue} \bf objectives} of being one which we agree with . \\
Replace & However , we agree with the {\color{sblue} \bf objectives} of the Ghilardotti report . \\
CDAlign & However , we agree with {\color{sblue} \bf objectives} of Fi{\color{sred} \bf Fiorella} Ghilardotti's report . \\
\cmidrule(lr){1-1} \cmidrule(lr){2-2}
Ours & We agree with the {\color{sblue} \bf objectives} of {\color{sred} \bf Fiorella} Ghilardotti's report , however .  \\
\bottomrule
\end{tabular}
\caption{Example translations of different lexically constrained NMT approaches.}
\label{tab:case-study}
\end{table*}

\section{Related Work}

\subsection{Lexically Constrained NMT}

One line of approaches to lexically constrained NMT focuses on designing advanced decoding algorithms~\cite{Hasler:2018:Lexical}. \citet{Hokamp:2017:Lexical} propose {\em grid beam search} (GBS), which enforces target constraints to appear in the output by enumerating constraints at each decoding step. The beam size required by GBS varies with the number of constraints. \citet{Post:2018:Lexical} propose {\em dynamic beam allocation} (DBA) to fix the problem of varying beam size for GBS, which is then extended by \citet{Hu:2019:Lexical} into VDBA that supports batched decoding.
There are also some other constrained decoding algorithms that leverage word alignments to impose constraints~\cite{Song:2020:Lexical,Chen:2021:Lexical}. Although the alignment-based decoding methods are faster than VDBA, they may be negatively affected by noisy alignments, resulting in low CSR. Recently, \citet{Susanto:2020:Lexical} adopt Levenshtein Transformer~\cite{Gu:2019:Levenshtein} to insert target constraints in a non-autoregressive manner, for which the constraints must be provided with the same order as that in the reference.

Another branch of studies proposes to edit the training data to induce constraints~\cite{Sennrich:2016:Lexical}. \citet{Song:2019:Lexical} directly replace source constraints with their target translations and \citet{Dinu:2019:Lexical} insert target constraints into the source sentence without removing source constraints. Similarly, \citet{Chen:2020:Lexical} propose to append target constraints after the source sentence.

In this work, we propose to integrate vectorized lexical constraints into NMT models. Our work is orthogonal to both constrained decoding and constraint-oriented data augmentation.
A similar work to us is that \citet{Li:2020:Lexical} propose to use continuous memory to store only the target constraint, which is then integrated into NMT models through the decoder self-attention. However, \citet{Li:2020:Lexical} did not exploit the correspondence between keys and values to model both source and target constraints.

\subsection{Controlled Text Generation} Recent years have witnessed rapid progress in controlled text generation. \citet{Dathathri:2020:Plug} propose to use the gradients of a discriminator to control a pre-trained language model to generate towards a specific topic. 
\citet{Liu:2021:DExperts} propose a decoding-time method that employs experts to control the generation of pre-trained language models.

We borrow the idea presented in \citet{Pascual:2021:PlugAndPlay} to insert a plug-in into the output layer. The difference between our plug-in network and \citet{Pascual:2021:PlugAndPlay} is that we use an input-dependent gate to control the effect of the plugged probability.

\section{Conclusion}

In this work, we propose to vectorize and integrate lexical constraints into NMT models. Our basic idea is to use the correspondence between keys and values in attention modules to model constraint pairs. Experiments show that our approach can outperform several representative baselines across four different translation directions. In the future, we plan to vectorize other attributes, such as the topic, the style, and the sentiment, to better control the generation of NMT models.

\section*{Acknowledgments}

This work is supported by the National Key R\&D Program of China (No. 2018YFB1005103), the National Natural Science Foundation of China (No. 61925601, No. 62006138), and the Tencent AI Lab Rhino-Bird Focused Research Program (No. JR202031). We sincerely thank Guanhua Chen and Chi Chen for their constructive advice on technical details, and all the reviewers for their valuable and insightful comments.

\bibliography{anthology,custom}
\bibliographystyle{acl_natbib}


\end{document}